\documentclass[information,article,submit,pdflatex,moreauthors]{Definitions/mdpi} 

\firstpage{1} 
\pubvolume{1}
\issuenum{1}
\articlenumber{0}
\pubyear{2023}
\copyrightyear{2023}
\datereceived{ } 
\daterevised{ } 
\dateaccepted{ } 
\datepublished{ } 
\hreflink{https://doi.org/} 

\usepackage{svg}
\usepackage{smartdiagram}  

\Title{Hybrid Classic-Quantum Computing for Staging of Invasive Ductal Carcinoma of Breast}

\TitleCitation{Hybrid Classic-Quantum Computing for Staging of Invasive Ductal Carcinoma of Breast}

\Author{Vicente Moret-Bonillo $^{1}$*\orcidA{}, Eduardo Mosqueira-Rey $^{1}$\orcidB{}, Samuel Magaz-Romero $^{1}$\orcidC{} and Diego Alvarez-Estevez $^{1}$\orcidD{}}

\AuthorNames{Vicente Moret-Bonillo, Eduardo Mosqueira-Rey, Samuel Magaz-Romero and Diego Alvarez-Estevez}

\AuthorCitation{Moret-Bonillo V., Mosqueira-Rey E.,Magaz-Romero S. and Alvarez-Estevez D.}

\address[1]{%
$^{1}$ \quad Universidade da Coruña, CITIC, Campus de Elviña, 15071 A Coruña, España; citic@citic-research.org}

\corres{Correspondence: vicente.moret@udc.es}

\abstract{Despite the great current relevance of Artificial Intelligence, and the extraordinary innovations that this discipline has brought to many fields -among which, without a doubt, medicine is found-, experts in medical applications of Artificial Intelligence are looking for new alternatives to solve problems for which current Artificial Intelligence programs do not provide with optimal solutions. For this, one promising option could be the use of the concepts and ideas of Quantum Mechanics, for the construction of quantum-based Artificial Intelligence systems. From a hybrid classical-quantum perspective, this article deals with the application of quantum computing techniques for the staging of Invasive Ductal Carcinoma of the breast. It includes: (1) a general explanation of a classical, and well-established, approach for medical reasoning, (2) a description of the clinical problem, (3) a conceptual model for staging invasive ductal carcinoma, (4) some basic notions about Quantum Rule-Based Systems, (5) a step-by-step explanation of the proposed approach for quantum staging of the invasive ductal carcinoma, and (6) the results obtained after running the quantum system on a significant number of use cases. A detailed discussion is also provided at the end of this paper.}

\keyword{Medical Reasoning; Invasive Ductal Carcinoma; Quantum Computing; Artificial Intelligence.} 

\begin{document}
\section{Introduction}

Despite the great current relevance of Artificial Intelligence (AI), and the extraordinary innovations that this discipline has brought to medicine, experts in medical applications of artificial intelligence are looking for new alternatives to solve problems for which current artificial intelligence programs do not provide with optimal solutions \cite{mintz2019introai,holzinger2019aimedicine}. For this, one promising option could be the use of the concepts and ideas of Quantum Mechanics (QM) and Quantum Computing (QC), for the construction of quantum-based AI systems.

Quantum Computing is based on the use of qubits instead of bits and leads to new logic gates that make it possible the construction of new quantum algorithms for the solution of a given task, theoretically improving complexity, efficiency, or energy consumption \cite{nielsen2010quantum}. In classical computing, a bit can only carry two values: 0 or 1. On the other hand, in Quantum Computing, the laws of Quantum Mechanics intervene, and the particle, the qubit, can be in superposition: it can be 0, it can be 1, and it can be 0 and 1 at the same time (two orthogonal states of a subatomic particle). This allows several operations to be performed simultaneously, depending on the number of qubits.
With conventional bits, if we have a three-bit register, there would be eight possible values, and the register could only hold one of those values. On the other hand, if we have a vector of three qubits, the particle can take on eight different values at the same time thanks to quantum superposition. Thus, a vector of three qubits would allow to carry out an operation on eight different states concurrently. As expected, the number of operations is exponential with respect to the number of qubits.

But this quantum advantage could mean more than just performing calculations faster. To give an example: would it be profitable for a quantum simulator -or a quantum computer- to provide the same result than a classic computer, in a comparable time, but with much greater energy savings?

In any case, and in the short term, it seems clear that Hybrid Classical-Quantum Systems (HCQS) are more realistic approaches than just pure quantum systems. In fact, we are dealing with problems in the NISQ (Noisy Intermediate-Scale Quantum) era of QC \cite{preskill2018nisq}. The term “Noisy” is because we don’t have enough qubits to spare for error correction. The term “Intermediate-Scale” is because the number of quantum bits is too small to compute sophisticated quantum algorithms but large enough to show quantum advantage or even supremacy. HCQS could potentially overcome the limitations of the NISQ era, as we intend to suggest in this paper.

This article deals with the application of quantum computing techniques for the staging of Invasive Ductal Carcinoma (IDC) of the breast. This article includes: (1) a general explanation of a classical, and well-established, approach for medical reasoning, (2) a description of the clinical problem, (3) a conceptual model for staging invasive ductal carcinoma, (4) some basic notions about Quantum Rule-Based Systems, (5) a step-by-step explanation of the proposed approach for quantum staging of the invasive ductal carcinoma, and (6) the results obtained after running the quantum system on a significant number of use cases. A detailed discussion is also provided at the end of this paper.

\section{Materials and Methods}

In this section we introduce the theoretical bases of the presented work, consisting of: (1) Categorical Reasoning in Medicine, (2) cancer staging and (3) Quantum Rule-Based Systems.

\subsection{Categorical Reasoning in Medicine}

In 1959, Ledley and Lusted published an article entitled "Reasoning Foundations of Medical Diagnosis" \cite{ledley1959reasoning}. In that article they analysed the inherent difficulty of the problem of medical diagnosis. The initial hypothesis of Ledley and Lusted is that the rational use of computers could greatly simplify the process of medical diagnosis, or at least, could provide an important resource for the efficient management of the relevant information to obtain a medical diagnosis.

However, at the time this important article was written, computers faced with a huge challenge, since their storage capabilities, and computing capabilities, were many orders of magnitude below of today’s current computers. This situation allows us to establish a first conclusion: Although the theoretical foundations of the approaches proposed by Ledley and Lusted were reasonably clear, the complexity of the knowledge representations associated with the proposed methods greatly limited the size of the problems that could be handled. 

In their approach, Ledley and Lusted identify three factors that play a key role in diagnostic tasks: (1) the medical knowledge that relates sets of symptoms with possible sets of diagnoses, (2) the sets of symptoms observed in a specific case, and (3) the sets of possible diagnoses that are consistent with the observed symptoms and with the knowledge of the physician who is analysing the case. In this context, it seems clear that this knowledge is mainly heuristic.

An important feature of the analysis of Ledley and Lusted is that they deal with the problem of medical diagnosis from the point of view of the differential interpretations that, basically, consists in eliminating situations of incompatibility between symptoms and possible diagnoses, given the knowledge domain.

This approach to the problem is very common in medicine, but from a computational perspective, and more specifically from the viewpoint of knowledge engineering and artificial intelligence this way to face the problem greatly complicates the knowledge acquisition task, a task that is always subjective, and that usually results in a computational model of apparently intelligent behaviour, but often incomplete, sometimes inefficient, and eventually unnatural. Let us explain the method proposed by Ledley and Lusted for the medical diagnosis task:

\begin{itemize}
    \item Let $n$ be the number of relevant specific symptoms in the domain.
    \item If a given patient has symptom $i$ then we say that $s_i = 1$
    \item If a given patient has not symptom $i$ then we say that $s_i = 0$ 
\end{itemize}

We are assuming a strictly categorical approach. Thus, in this domain, if $n$ is the total number of specific symptoms, we have $2^n$ possible combinations of symptoms. To avoid ambiguity, we adopt the binary numbering system. Thus, if $n = 3$ then we will have $2^3 = 8$ possible associations of symptoms: $S0 \ldots S7$ (see Table \ref{tab:symptoms}).

\begin{table}[H] 
\caption{The eight possible associations (upper case letters) that can be built depending on the values of three different symptoms (lower case letters).\label{tab:symptoms}}
\newcolumntype{C}{>{\centering\arraybackslash}X}
\begin{tabularx}{\textwidth}{CCCCCCCCC}
\toprule
  & $\mathbf{S_0}$ & $\mathbf{S_1}$ & $\mathbf{S_2}$ & $\mathbf{S_3}$ & $\mathbf{S_4}$ & $\mathbf{S_5}$ & $\mathbf{S_6}$ & $\mathbf{S_7}$ \\
\midrule
$\mathbf{s_1}$ & 0 & 0 & 0 & 0 & 1 & 1 & 1 & 1 \\
$\mathbf{s_2}$ & 0 & 0 & 1 & 1 & 0 & 0 & 1 & 1 \\
$\mathbf{s_3}$ & 0 & 1 & 0 & 1 & 0 & 1 & 0 & 1 \\
\bottomrule
\end{tabularx}
\end{table}

From this strictly categorical perspective, in a particular case, and in terms of associations, the symptoms of a patient are represented by one and only one of the associations of individual symptoms. We say that the set of associations of symptoms, $S = \{S_0, S_1, \ldots\}$ is complete and exhaustive, and its elements are mutually exclusive. The same statement is true if we talk about diagnoses.

Let us now consider that we have two possible symptoms $(S = \{s_1, s_2\})$ and two possible diagnoses $(D = \{d_1, d_2\})$. If we express this in terms of associations, we have:

\begin{equation*}
    S = \{S_0, S_1, S_2, S_3\} \qquad D = \{D_0, D_1, D_2, D_3\}
\end{equation*}

At this point, it seems reasonable to assume that there is a relationship between relevant symptoms and possible diagnoses. In this context, and working with the relevant associations of symptoms and diagnoses, we can build what Ledley and Lusted call Expanded Logic Base (ELB) as the cartesian product $S \times D$ (see Table \ref{tab:elb}).

\begin{table}[H] 
\caption{Expanded Logic Base (ELB) built from the cartesian product of the set of associations of symptoms and the set of associations of diagnoses (S × D). \label{tab:elb}}
\newcolumntype{C}{>{\centering\arraybackslash}X}
\begin{tabularx}{\textwidth}{CCCCCCCCCCCCCCCCC}
\toprule
  & $\mathbf{S_0}$ & $\mathbf{S_1}$ & $\mathbf{S_2}$ & $\mathbf{S_3}$ & $\mathbf{S_0}$ & $\mathbf{S_1}$ & $\mathbf{S_2}$ & $\mathbf{S_3}$ & $\mathbf{S_0}$ & $\mathbf{S_1}$ & $\mathbf{S_2}$ & $\mathbf{S_3}$ & $\mathbf{S_0}$ & $\mathbf{S_1}$ & $\mathbf{S_2}$ & $\mathbf{S_3}$ \\
\midrule
$\mathbf{s_1}$ & 0 & 0 & 1 & 1 & 0 & 0 & 1 & 1 & 0 & 0 & 1 & 1 & 0 & 0 & 1 & 1 \\
$\mathbf{s_2}$ & 0 & 1 & 0 & 1 & 0 & 1 & 0 & 1 & 0 & 1 & 0 & 1 & 0 & 1 & 0 & 1\\
$\mathbf{d_1}$ & 0 & 0 & 0 & 0 & 0 & 0 & 0 & 0 & 1 & 1 & 1 & 1 & 1 & 1 & 1 & 1 \\
$\mathbf{d_2}$ & 0 & 0 & 0 & 0 & 1 & 1 & 1 & 1 & 0 & 0 & 0 & 0 & 1 & 1 & 1 & 1 \\
\midrule
 & \multicolumn{4}{c}{$\mathbf{D_0}$} & \multicolumn{4}{c}{$\mathbf{D_1}$} & \multicolumn{4}{c}{$\mathbf{D_2}$} & \multicolumn{4}{c}{$\mathbf{D_3}$} \\
\bottomrule
\end{tabularx}
\end{table}

Note that in ELB are absolutely all possible combinations of symptoms and diagnoses. Some of such combinations will be out of clinical sense and, therefore, they must be eliminated. In this regard, Ledley and Lusted propose the use of the knowledge domain to rule out absurd combinations. For this purpose, they face the problem from the differential diagnosis perspective. This involves removing from ELB those associations that are incompatible with the medical knowledge. We will try to explain, following the example proposed by Ledley and Lusted, how this works. Suppose that, after knowledge acquisition, we have been able to identify the following rules:

\begin{itemize}
    \item[R1:] \textit{If the patient has symptoms, then there must be one or more associated diagnosis.}
    
    (Note that this rule does not preclude asymptomatic problems, only it states that if there are symptoms is because there are problems)
    \item[R2:] \textit{If the patient has the disease $d_2$, then he or she must present the symptom $s_1$.}
    \item[R3:] \textit{If the patient has the disease $d_1$, but does not have the disease $d_2$, then he or she must present the symptom $s_2$.}
    \item[R4:] \textit{If the patient has not the disease $d_1$, but he or she has the disease $d_2$ then symptom $s_2$ must be absent.}
\end{itemize}

Now, following the differential approach, we must discard those associations that are in ELB, but that also are incompatible with the rules of our knowledge domain. The result is a Reduced Logic Base (RLB) containing only those associations symptom-diagnosis that are possible given the available knowledge. In this case:

\begin{equation*}
    RLB = \{S_0D_0, S_1D_2, S_2D_1, S_2D_3, S_3D_2, S_3D_3\}
\end{equation*}

Let us now illustrate these ideas with two hypothetical clinical cases:

\begin{enumerate}
    \item A patient goes to see the doctor presenting the symptoms $(s_1 = 0; s_2 = 1)$. These symptoms respond to the association of symptoms $S_1$, which is associated, in RLB, with the diagnostic complex $D_2 = (d_1 = 1; d_2 = 0)$. In this case we can say that the patient has the disease $d_1$, but we can rule out disease $d_2$.
    \item Assume now a different case in which the symptoms are $(s1 = 1; s2 = 0)$. In terms of associations, this is the complex of symptoms $S_2$, which is associated in RLB with $D_1$ and with $D_3$. But $D_1 = (d_1 = 0; d_2 = 1)$ and, on the other hand, $D_3 = (d_1 = 1; d_2 = 1)$. In this case, we can say that the patient has the disease $d_2$ with certainty, but we should not rule out the pathology $d_1$ because there is evidence for and against among which we cannot discriminate. 
\end{enumerate}

The conclusion is obvious: no matter how much categorical the procedure is -and this procedure is strictly categorical- uncertainty is inherent to human reasoning, and appears spontaneously and naturally. This fact is very important, and may be due to one, or all, of the following factors: (1) the nature of heuristic knowledge, (2) the inaccurate nature of knowledge, (3) the lack of knowledge, or (4) the subjectivity in the interpretation of the information.

\subsection{Description of the Clinical Problem}

Invasive Ductal Carcinoma (IDC), sometimes referred to as infiltrating ductal carcinoma, is the most common type of breast cancer. About 80\% of all cases of breast cancer are invasive ductal carcinomas.

Staging is the process used to estimate the extent of invasive ductal carcinoma spread from its original location. The stage of invasive ductal carcinoma is described on a scale from stage I (the earliest stage) to stage IV (the most advanced stage) \cite{amin2017staging}. 

Stage I describes invasive breast cancer (cancer cells take in or invade the normal breast tissue around them). Stage I is divided into subcategories, known as I-A and I-B. Stage I-A describes invasive breast cancer in which the tumour is up to 2 cm and no lymph nodes are affected. Stage I-B describes invasive breast cancer in which: (1) there is no tumour in the breast, (2) small groups of cancer cells greater than 0.2 mm but less than 2 mm are observed in the lymph nodes, or (3) there is a breast tumour smaller than 2 cm and small groups of cancer cells larger than 0.2 mm but smaller than 2 mm in the lymph nodes.

Stage II is divided into subcategories II-A and II-B. Stage II-A describes invasive breast cancer in which: (1) there is no tumour in the breast, but cancer cells are found in 1-3 axillary lymph nodes under the arm or in lymph nodes near the breastbone, or (3) the tumour is 2 cm or smaller and has spread to the axillary lymph nodes, or (4) the tumour is 2 to 5 cm and has not spread to the axillary lymph nodes. Stage II-B describes invasive breast cancer in which: (1) the tumour is between 2 and 5 cm, and small groups of cancer cells larger than 0.2 mm but smaller than 2 mm are seen in the lymph nodes, or (2) the tumour is 2 to 5 cm, and the cancer has spread to 1-3 axillary lymph nodes or lymph nodes near the breastbone, or (3) the tumour is larger than 5 cm but has not spread to the axillary lymph nodes.

Stage III is divided into subcategories III-A, III-B, and III-C.  Stage III-A describes invasive breast cancer in which: (1) the tumour may be any size, and cancer was found in 4-9 axillary lymph nodes or lymph nodes near the breastbone, or (2) the tumour is larger than 5 cm, and small clusters of cancer cells larger than 0.2 mm but smaller than 2 mm are seen in the lymph nodes, or (3) the tumour is larger than 5 cm, and the cancer has spread to 1-3 axillary lymph nodes or lymph nodes near the breastbone. Stage III-B describes invasive breast cancer in which the tumour is indefinite in size and has spread to the chest wall or skin of the breast. Stage III-C describes invasive breast cancer in which: (1) there may be no evidence of disease in the breast or, (2) if a tumour is present, it may be any size and may have spread to the chest wall or skin of the breast and cancer has spread to 10 or more axillary lymph nodes, or (3) cancer has spread to lymph nodes above or below the collarbone or cancer has spread to axillary lymph nodes or lymph nodes near the breastbone.Stage IV describes invasive breast cancer that has spread beyond the breast and surrounding lymph nodes to other organs in the body, such as the lungs, distant lymph nodes, skin, bones, liver, and brain.

\subsection{A Model for the Staging of IDC}

Knowledge engineering is a sub-field of artificial intelligence whose purpose is the design and development of expert systems. This is supported by instructional methodologies, trying to represent the human knowledge and reasoning in a certain domain, within an artificial system. The work of knowledge engineers consists of extracting the knowledge of human experts, and in coding said knowledge so that it can be processed by a computer system. The problem is that the knowledge engineer is not an expert in the field that tries to model, while the expert in the subject has no experience modelling his/her knowledge (based on heuristics) in a way that can be represented generically in the computer system. 

Knowledge engineering encompasses the scientists, technology and methodology required to process knowledge. The goal is extract, articulate, and computerize knowledge from an expert. The result is a knowledge model, ready to be implemented and tested. In this case, we use the method provided by the TNM staging system \cite{giuliano2018breastcancer} to describe the amount and spread of cancer in a patient’s body, where:

\begin{itemize}
    \item T describes the size of the tumour and any spread of cancer into nearby tissue. 
    \item N describes spread of cancer to nearby lymph nodes. 
    \item M describes metastasis (spread of cancer to other parts of the body). 
\end{itemize}

\begin{figure}[H]
\includegraphics[width=\textwidth]{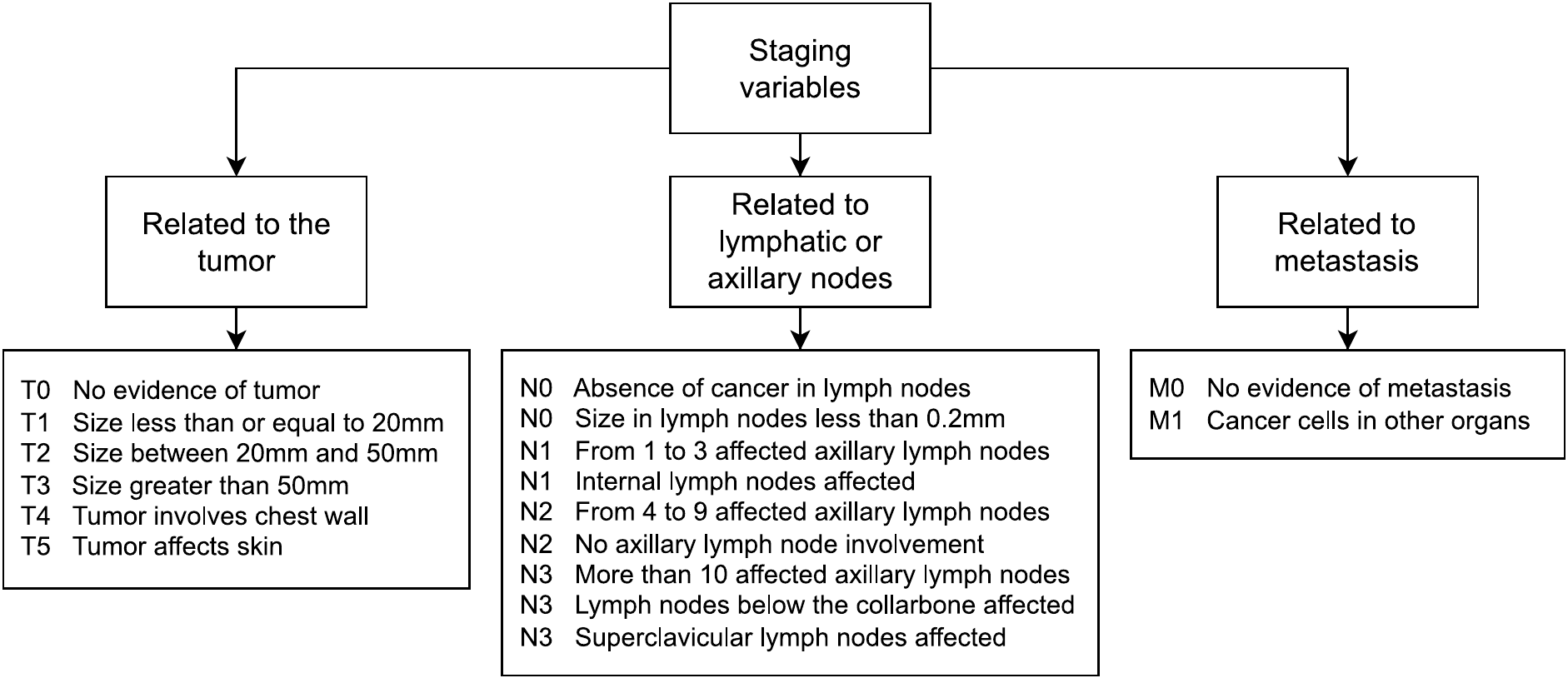}
\caption{Variables or symptoms that are considered in the knowledge model for IDC staging. \label{fig:staging_variables}}
\end{figure}

Each of these categories is furtherly detailed with numbers or letters, differentiating the severeness of the case. We use a reduced version of this system, only employing the numbers notation and using X/Y to denote that any number would fit. Figure \ref{fig:staging_variables} illustrates the variables we are considering in our knowledge model for staging IDC, and Table \ref{tab:tnm_idc} shows the correspondence between TNM classifications and IDC stages.

\unskip

\begin{table}[H] 
\caption{Invasive Ductal Carcinoma stages according with TNM classification system.\label{tab:tnm_idc}}
\newcolumntype{C}{>{\centering\arraybackslash}X}
\begin{tabularx}{\textwidth}{Cl}
\toprule
\textbf{IDC Stage}	& \textbf{Compatible TNM classification}\\
\midrule
I-A   & T1 N0 M0 \\
I-B	  & T0 N1 M0 / T1 N1 M0 \\
II-A  & T0 N1 M0 / T1 N1 M0 / T2 N0 M0 \\
II-B  & T2 N1 M0 / T3 N0 M0 \\
III-A & T0 N2 M0 / T1 N2 M0 / T2 N0 M0 / T3 N2 M0 / T3 N1 M0 \\
III-B & T4 N0 M0 / T4 N1 M0 / T4 N2 M0 \\
III-C & TX N3 M0 \\
IV    & TX NY M1  \\
\bottomrule
\end{tabularx}
\end{table}

\subsection{Quantum Rule-Based Systems}

Quantum Rule-Based Systems (QRBS) are defined as those Rule-Based Systems (RBS) that use the formalism of Quantum Computing (QC) for representing knowledge and for making inferences \cite{moret2018emerging,moret2021uncertainty,moret2022quantum}. Let us consider the following set of rules:

\begin{itemize}
    \item[R1:] IF A and B THEN X 
    \item[R2:] IF X or C THEN Y 
    \item[R3:] IF Y and (D or E) THEN R 
\end{itemize}

\begin{figure}[H]
\includegraphics[width=0.45\textwidth]{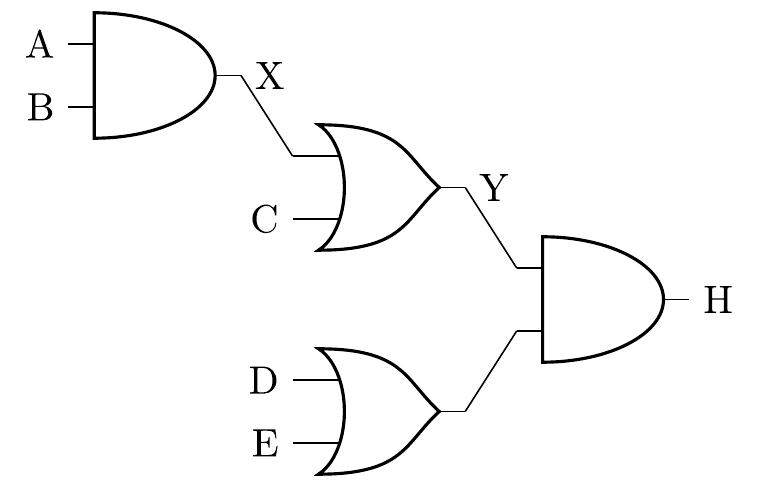}
\caption{Classical representation of the inferential circuit of the example. \label{fig:inferential_circuit}}
\end{figure}

In conventional RBS, any categorical rule can be represented by the logical operators \textit{\{and, or, not\}} that relate statements that are always true  (i.e. assuming no uncertainty nor imprecision whatsoever). Thus, rule R1 should be interpreted as follows: If statement A is true, and statement B is true, then we can conclude without uncertainty that statement X is true. The three previously defined rules can be represented classically by means of the inferential circuit of Figure \ref{fig:inferential_circuit}.

\begin{figure}[H]
\includegraphics[width=0.4\textwidth]{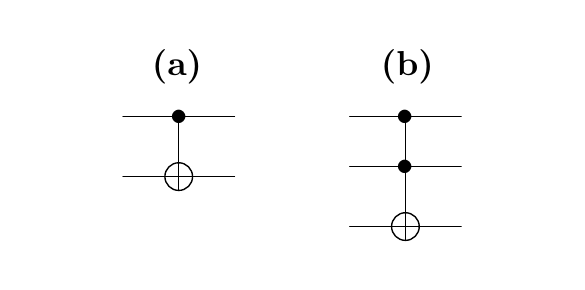}
\caption{Quantum representation of reversible logical gates: (\textbf{a}) is the CN gate, (\textbf{b}) is the CCN gate. In the figure, lines with a black dot ($\bullet$) are control lines and the symbol ($\oplus$) is a quantum-not operator. \label{fig:quantum_cnot_ccnot}}
\end{figure}

However, if we choose the formalism of QC, we need reversible quantum gates to represent the previous inferential circuit. In order to translate the classical RBS inferential circuit of Figure \ref{fig:inferential_circuit} to a quantum environment, we need to rewrite the logical operators \textit{{and, or, not}} following the restrictions imposed by quantum computing. We use the quantum circuit approach, for which we need  reversible quantum gates such as the CN (controlled-not gate) and the CCN gate (controlled-controlled-not gate). Figure \ref{fig:quantum_cnot_ccnot} illustrates how these reversible quantum gates looks like \cite{yanosfky2008computerscientists}. 

\begin{figure}[H]
\includegraphics[width=1\textwidth]{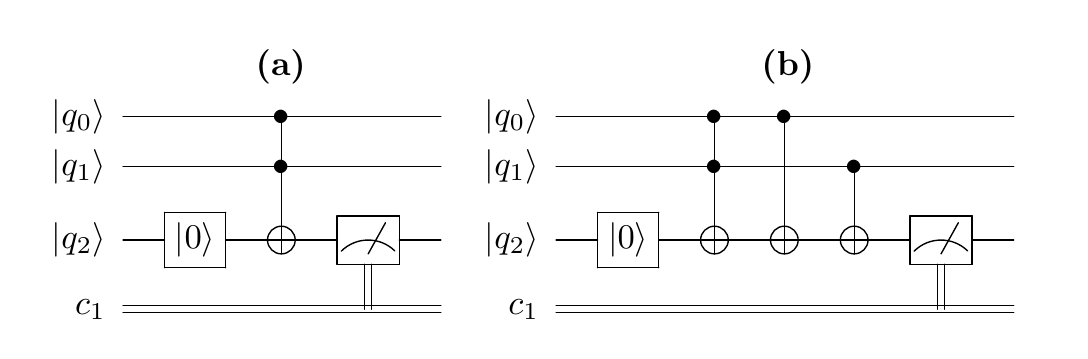}
\caption{Quantum representation of reversible logical operators: (\textbf{a}) is the \textit{{quantum-and}}, (\textbf{b}) is the \textit{{quantum-or}}. Quantum lines $q_0$ and $q_1$ are the inputs, that can be in state 0 or in state 1. Quantum line $q_2$ is the output, that -initially- must be always in state 0. After the operation has been done, we measure $q_2$ and the result of the measurement is retrieved from the conventional-classic line $c_1$. \label{fig:quantum_and_or}}
\end{figure}

These two quantum gates allow us the construction of the corresponding \textit{{quantum-and}} and \textit{{quantum-or}} operators (see Figure \ref{fig:quantum_and_or}).

\begin{figure}[H]
\includegraphics[width=0.75\textwidth]{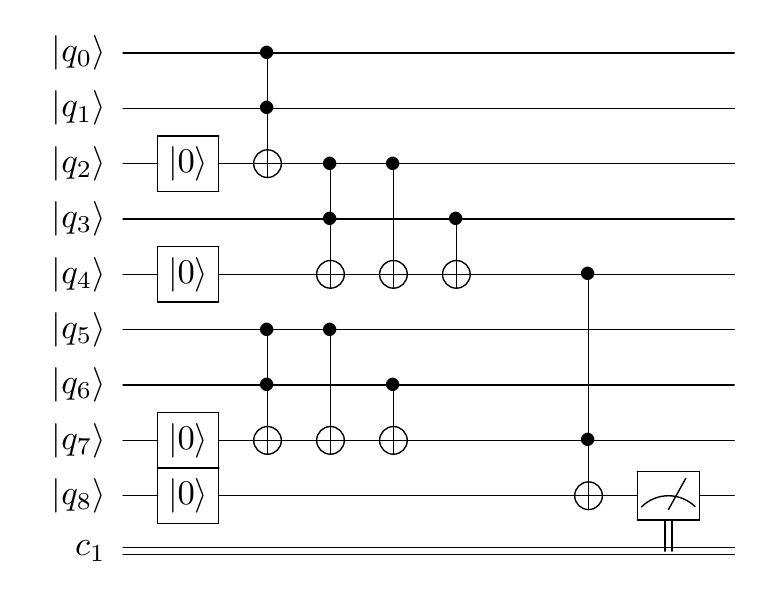}
\caption{Quantum representation of the inferential circuit of Figure \ref{fig:inferential_circuit}. \label{fig:quantum_inferential}}
\end{figure}

Now, we can represent the above-mentioned inferential circuit (see Figure \ref{fig:inferential_circuit}) in a quantum manner (see Figure \ref{fig:quantum_inferential}).

\subsection{Quantum System Design for Breast IDC Staging}

\begin{figure}[H]
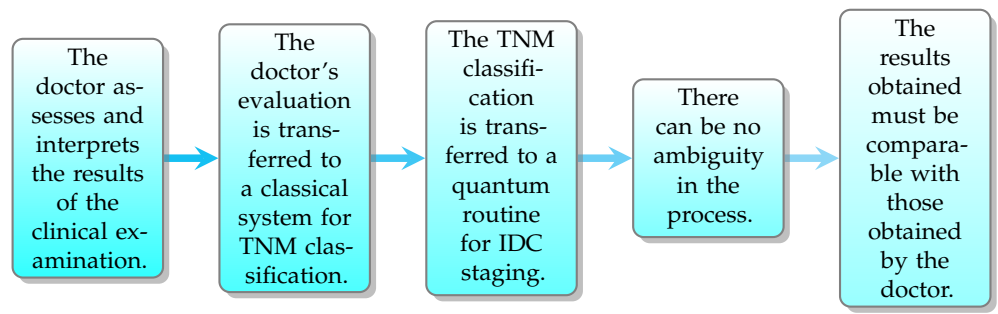

\smartdiagramset{
    back arrow disabled=true,
    set color list={cyan!80, cyan!70, cyan!60, cyan!50, cyan!40}
}
\smartdiagram[flow diagram:horizontal]{
    The doctor assesses and interprets the results of the clinical examination., 
    The doctor's evaluation is transferred to a classical system for TNM classification., 
    The TNM classification is transferred to a quantum routine for IDC staging., 
    There can be no ambiguity in the process., 
    The results obtained must be comparable with those obtained by the doctor.
}  
\caption{Framework for Beast Invasive Ductal Carcinoma Staging.\label{fig:framework}}
\end{figure}

We have to consider the following set of assumptions and domain restrictions, which will condition the design of the proposed classic-quantum hybrid solution for breast IDC staging: (1) the participation and collaboration of clinical experts is essential, (2) the magnitude of the problem is limited by the small number of qubits from the NISQ era of Quantum Computing, which requires us to optimize the overall process by reducing as much as possible the number of qubits required , (3) the final solution must be neither pure quantum nor pure classical. It must be a hybrid solution due to the interactions with the experts , and (4) the results obtained must be comparable to what a human expert would obtain. This framework can be represented as shown in Figure \ref{fig:framework}.

Next, and considering the knowledge model detailed in Figure \ref{fig:staging_variables} and Table \ref{tab:tnm_idc} , we will focus on the description of the quantum routine for IDC staging. If we consider the TNM staging system, and in reference to the categorial reasoning notation introduced in previous sections, notice that for this case \textit{S = \{T0, T1, T2, T3, T4, T5, N0, N1, N2, N3, M0, M1\}} and \textit{D = \{I-A, I-B, II-A, II-B, III-A, III-B, III-C, IV\}}. Applying now the corresponding medical reasoning model, only the complexes shown in Table \ref{tab:tnm_qubits} are relevant.  For each of these relevant complexes, which can be considered as vectors, a qubit will be associated to it, which will be the input to the quantum subroutine . Since a given patient cannot be in two different TNM states at the same time, the quantum subroutine is activated only when one and only one of the input qubits is in state 1, and all the others are in state 0.

\begin{table}[H] 
\caption{Relevant TNM states and corresponding input qubits to the quantum system.\label{tab:tnm_qubits}}
\newcolumntype{C}{>{\centering\arraybackslash}X}
\begin{tabularx}{\textwidth}{CCCCCCCCCCCCCCCC}
\toprule
\multicolumn{1}{c}{\multirow{2}{*}{TNM}} & \multicolumn{15}{c}{Input qubit}\\ \cmidrule{2-16}
   & q0 & q1 & q2 & q3 & q4 & q5 & q6 & q7 & q8 & q9 & q10 & q11 & q12 & q13 & q14 \\ \midrule
T0 & X  & X  &    &    &    &    &    &    &    &    &     &     &     &     &     \\ \midrule
T1 &    &    & X  & X  & X  &    &    &    &    &    &     &     &     &     &     \\ \midrule
T2 &    &    &    &    &    & X  & X  &    &    &    &     &     &     &     &     \\ \midrule
T3 &    &    &    &    &    &    &    & X  & X  & X  &     &     &     &     &     \\ \midrule
T4 &    &    &    &    &    &    &    &    &    &    & X   & X   & X   &     &     \\ \midrule
N0 &    &    & X  &    &    & X  &    & X  &    &    & X   &     &     &     &     \\ \midrule
N1 & X  &    &    & X  &    &    & X  &    & X  &    &     & X   &     &     &     \\ \midrule
N2 &    & X  &    &    & X  &    &    &    &    & X  &     &     & X   &     &     \\ \midrule
N3 &    &    &    &    &    &    &    &    &    &    &     &     &     & X   &     \\ \midrule
M0 & X  & X  & X  & X  & X  & X  & X  & X  & X  & X  & X   & X   & X   &     &     \\ \midrule
M1 &    &    &    &    &    &    &    &    &    &    &     &     &     &     & X   \\ 
\bottomrule
\end{tabularx}
\end{table}

On the other hand, we must design our quantum routine in such a way that -after the measurement process- we obtain a string of conventional bits that can be directly associated with the stage of the cancer. Also, since there are eight possible stages of cancer, we need an eight-bit string in the output. Table \ref{tab:output} illustrates the matching between the output bit string and the different stages of cancer.

\begin{table}[H] 
\caption{Correspondence between the output bits and IDC stage.\label{tab:output}}
\newcolumntype{C}{>{\centering\arraybackslash}X}
\begin{tabularx}{\textwidth}{CCCCCCCCC}
\toprule
\textbf{Output Bit}	& c7 & c6 & c5 & c4 & c3 & c2 & c1 & c0 \\
\midrule
\textbf{IDC Stage}	& IV & III-C & III-B & III-A & II-B & II-A & I-B & I-A \\
\bottomrule
\end{tabularx}
\end{table}

\begin{figure}[H]
\begin{adjustwidth}{-\extralength}{0cm}
\includegraphics[width=1.35\textwidth]{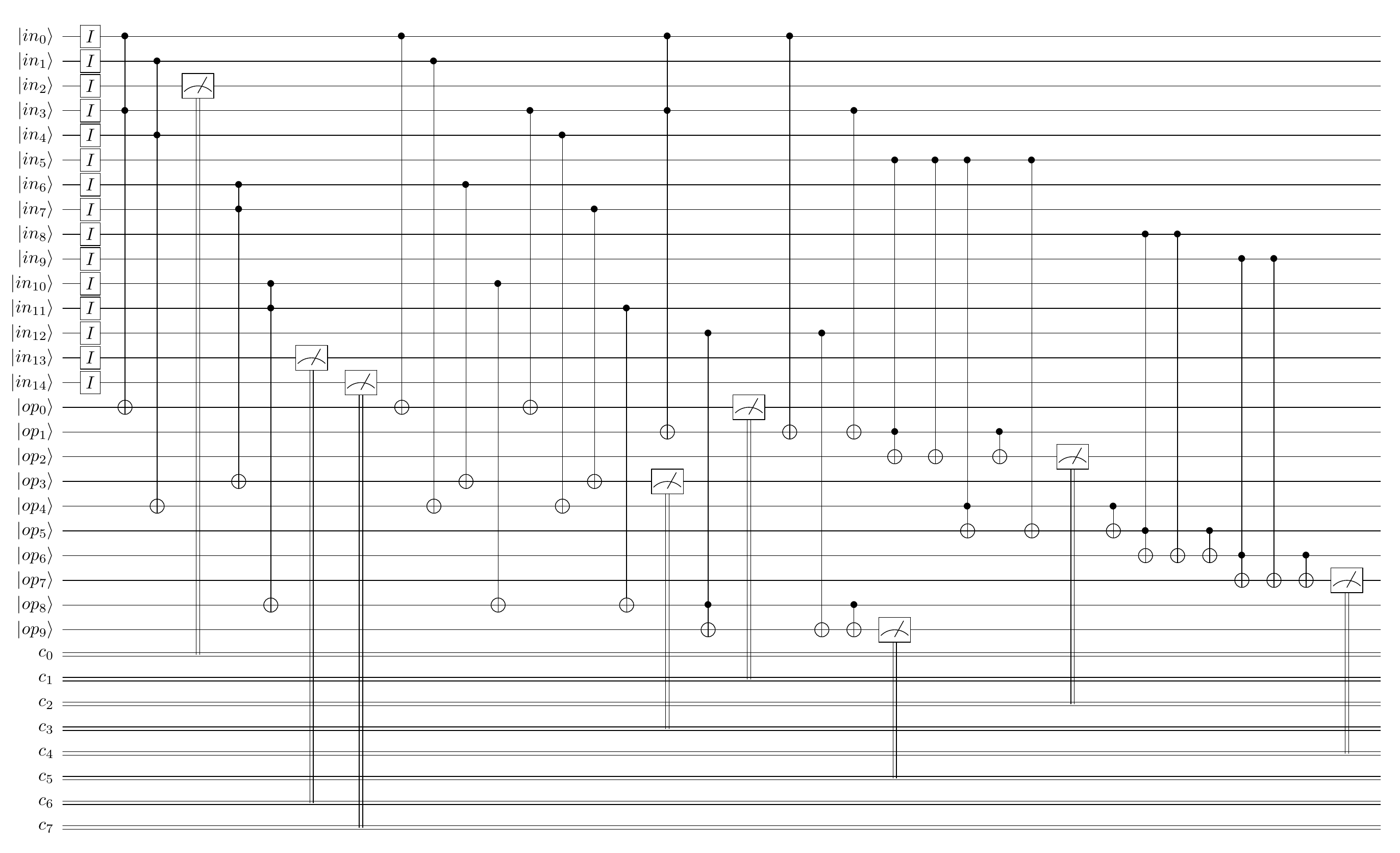}
\caption{Architecture of the Quantum Subroutine for Breast Invasive Ductal Carcinoma Staging. \label{fig:quantum_idc}}
\end{adjustwidth}   
\end{figure}

An added problem is that, according to what is stated in Table \ref{tab:tnm_idc}, in each specific case the same TNM classification may be compatible with several different stages of cancer. This fact must be considered in the design of our quantum routine. This implies defining and implementing a set of quantum rules (relating the TNM classification input to the IDC staging output) and configuring the inferential circuits that represent the different possible cases. For this we need to use an extra set of qubits on which the logical operations imposed by the quantum rules are carried out. In our case, we are going to need 10 extra qubits to solve the defined inferential circuits . Thus, for the quantum resolution of the IDC staging, we need: (a) 15 input qubits to set up the problem, (b) 10 extra qubits to perform the logical operations imposed by the inferential circuits, and (c) a string of 8 conventional bits, on which -after the corresponding measurements- the results obtained by the quantum subroutine will be written. The resulting architecture of the quantum subroutine is illustrated in Figure \ref{fig:quantum_idc}.

The quantum circuit works as follows: note that, conventionally, the initial state of any qubit is always 0. To activate it, one -and only one- of the input qubits must be ``negated'' (set to state 1) and the quantum system begins to solve the corresponding quantum circuits to finally obtain the string of conventional bits, which allows us to identify the IDC stages that are compatible with the input information. For this quantum subroutine, there are 15 different inputs that can activate it, one for each individual qubit.

\section{Experiments and results}

A preliminary validation of the proposed approach was carried out, in which: (1) the TNM classification obtained by the classical subroutine is transferred to the quantum subroutine as an \textit{Activated Qubit}, (2) the corresponding qubit, initially in state 0, is activated and changes its state to 1, (3) the quantum subroutine starts to work, (4) the cancer stages that are compatible with the TNM classification are obtained, and (5) the final solution is a chain of bits.

Table \ref{tab:results} shows the results of the experiment, in which the inferred cancer stages are also compared with the expected cancer stages. The experiment was carried out with the IBM quantum simulator.

\begin{table}[H] 
\caption{Results of the experiment carried out.\label{tab:results}}
\newcolumntype{C}{>{\centering\arraybackslash}X}
\begin{tabularx}{\textwidth}{CCCCC}
\toprule
\textbf{TNM Classification} & \textbf{Activated Qubit} & \textbf{Output Bits} & \textbf{IDC Stage Calculated} & \textbf{IDC Stage Expected}\\
\midrule
T0 N1 M0 & q0  & 00000110 & I-B or II-A	    & I-B or II-A \\
T0 N2 M0 & q1  & 00010000 & III-A	        & III-A \\
T1 N0 M0 & q2  & 00000001 & I-A	            & I-A \\
T1 N1 M0 & q3  & 00000110 & I-B or II-A	    & I-B or II-A \\
T1 N2 M0 & q4  & 00010000 & III-A	        & III-A \\
T2 N0 M0 & q5  & 00010100 & III-A or II-A	& III-A or II-A \\
T2 N1 M0 & q6  & 00001000 & II-B	        & II-B \\
T3 N0 M0 & q7  & 00001000 & II-B	        & II-B \\
T3 N1 M0 & q8  & 00010000 & III-A	        & III-A \\
T3 N2 M0 & q9  & 00010000 & III-A	        & III-A \\
T4 N0 M0 & q10 & 00100000 & III-B	        & III-B \\
T4 N1 M0 & q11 & 00100000 & III-B	        & III-B \\
T4 N2 M0 & q12 & 00100000 & III-B	        & III-B \\
TX N3 M0 & q13 & 01000000 & III-C	        & III-C \\
TX NY M1 & q14 & 10000000 & IV	            & IV \\
\bottomrule
\end{tabularx}
\end{table}

\section{Discussion  and conclusions}

For most types of cancer, doctors need to know how much cancer there is and where it is. This helps determine the most appropriate treatment options. For example, the best treatment for early-stage cancer may be surgery or radiation therapy, while advanced-stage cancer may require treatment that involves all parts of the body, such as chemotherapy, targeted drug therapy, or immunotherapy. The stage of a cancer can also be useful in helping to predict the likely progression of the cancer, as well as the likelihood that any treatment will be effective. Although each person's situation is unique, cancers of the same type and stage often have different prognoses. 

In this context, Biomedical Informatics is a powerful aid for decision-making in the field of medicine. In this sense, computing has provided many programs whose usefulness is evident, but insufficient. Now a new computing paradigm is emerging -quantum computing- and almost immediately quantum computing applied to medicine is born. In principle, it is expected to surpass classical computing in its field of action. The relationship between QC and health becomes increasingly close and interconnected. Thanks to quantum computing processes, problems that would take a lifetime to solve with all conventional supercomputers running in parallel, will be solved in seconds, hours, or days. However, in the opinion of the authors of this work, quantum computing will not replace classical computing, but will improve and complement it. Even for problems that a quantum computer can solve best, conventional computers will be needed, because data input and output will still be done in the traditional way. The next great revolution in computing will take place when quantum computing is combined with classical computing, and with artificial intelligence. The synergies emanating from this triad will guide the future of computing. Following this line of reasoning, in this article we have presented a hybrid classical-quantum approach to provide an alternative solution to the IDC staging problem. The results obtained are promising. However, one must also keep in mind that we are at the dawn of QC, and that there is still a lot to do.

\authorcontributions{Conceptualization, V. M-B.; methodology, V. M-B.; software, S. M-R.; validation, V. M-B. and S. M-R.; formal analysis, V. M-B.; investigation, V. M-B.; resources, S. M-R.; writing---original draft preparation, V. M-B. and S. M-R.; writing---review and editing, E. M-R. and D. A-E.; supervision, V. M-B.; project administration, V. M-B.; funding acquisition, V. M-B. All authors have read and agreed to the published version of the manuscript.}


\funding{This work was supported by the European Union’s Horizon 2020 research and innovation programme under project NEASQC (grant agreement No. 951821), by the Xunta de Galicia (ED431C 2022/44 and ED431H 2020/10) with the European Union ERDF funds, and by Spanish AIE (PID2019-107194GB-I00 / AEI / 10.13039/501100011033). We wish to acknowledge the support received from the Centro de Investigación de Galicia ``CITIC'', funded by Xunta de Galicia under collaboration agreement between Consellería de Cultura, Educación, Formación Profesional e Universidades and Galician Universities for reinforcement of research centres of Sistema Universitario de Galicia (CIGUS).}

\conflictsofinterest{The authors declare no conflict of interest. The funders had no role in the design of the study; in the collection, analyses, or interpretation of data; in the writing of the manuscript; or in the decision to publish the results.} 

\abbreviations{Abbreviations}{
The following abbreviations are used in this manuscript:\\

\noindent 
\begin{tabular}{@{}ll}
AI & Artificial Intelligence \\
CN & Controlled-Not \\
CCN & Controlled-Controlled-Not \\
HQCS & Hybrid Classical-Quantum Systems \\
IDC & Invaise Ductal Carcinoma \\
NISQ & Noisy Intermediate-Scale Quantum \\
QC & Quantum Computing \\
QM & Quantum Mechanics \\
QRBS & Quantum Rule-Based System \\
RBS & Rule-Based System \\
\end{tabular}
}

\begin{adjustwidth}{-\extralength}{0cm}

\reftitle{References}


\bibliography{bibliography}

\PublishersNote{}
\end{adjustwidth}
\end{document}